\PassOptionsToPackage{table}{xcolor}
\documentclass[11pt]{article}

\usepackage{acl}
\usepackage{times}
\usepackage{latexsym}

\usepackage[T1]{fontenc}
\usepackage{tabularx}
\usepackage[utf8]{inputenc}
\usepackage{afterpage}
\usepackage{microtype}

\usepackage{inconsolata}

\usepackage{graphicx}

\usepackage{booktabs}
\usepackage{algorithm}
\usepackage{algorithmic}
\usepackage{amsmath}

\usepackage{graphicx} 

\usepackage{amssymb}
\usepackage{pifont}

\usepackage{siunitx}
\usepackage{multirow}
\usepackage{longtable}
\usepackage{hyperref}
\usepackage{booktabs} 
\usepackage{graphicx}  
\usepackage{multirow}  
\usepackage{makecell}  
\usepackage{booktabs}  
\usepackage{adjustbox}
\author{
\bf Fanyi Yang\textsuperscript{1}\thanks{Work done during internship at Microsoft.},
\bf Jianfeng Liu\textsuperscript{2}\thanks{Corresponding Author.},
\bf Xin Zhang\textsuperscript{2},
\bf Haoyu Liu\textsuperscript{2},
\bf Xixin Cao\textsuperscript{1}, \\
\bf Yuefeng Zhan\textsuperscript{2}, 
\bf Hao Sun\textsuperscript{2},
\bf Weiwei Deng\textsuperscript{2},
\bf Feng Sun\textsuperscript{2},
\bf Qi Zhang\textsuperscript{2} \\
\\
\small{\textsuperscript{1}Peking University \quad \textsuperscript{2}Microsoft Corporation} \\
\small{yangfanyi@stu.pku.edu.cn, cxx@ss.pku.edu.cn} \\
\small{\{jianfengliu, xinzhang3, yuefzh, hasun, dedeng, sunfeng, qizhang\}@microsoft.com} \\
\small{implhy@gmail.com}
}

\begin{document} 
\title{MAIN: Mutual Alignment Is Necessary for instruction tuning }
\maketitle

\begin{abstract}
Instruction tuning has empowered large language models (LLMs) to achieve remarkable performance, yet its success heavily depends on the availability of large-scale, high-quality instruction-response pairs. To meet this demand, various methods have been developed to synthesize data at scale. However, current methods for scaling up data generation often overlook a crucial aspect: the alignment between instructions and responses. We hypothesize that the quality of instruction-response pairs is determined not by the individual quality of each component, but by the degree of mutual alignment. To address this, we propose a Mutual Alignment Framework (MAIN) which enforces coherence between instructions and responses through mutual constraints. We demonstrate that MAIN generalizes well across model architectures and sizes, achieving state-of-the-art performance on LLaMA, Mistral, and Qwen models across diverse benchmarks. This work underscores the critical role of instruction-response alignment in enabling generalizable and high-quality instruction tuning for LLMs. All code is available from \href{https://github.com/stamina1121/MAIN}{our repository}.

\end{abstract}

\begin{figure}[t]
    \centering
        \includegraphics[width=1.0\linewidth]{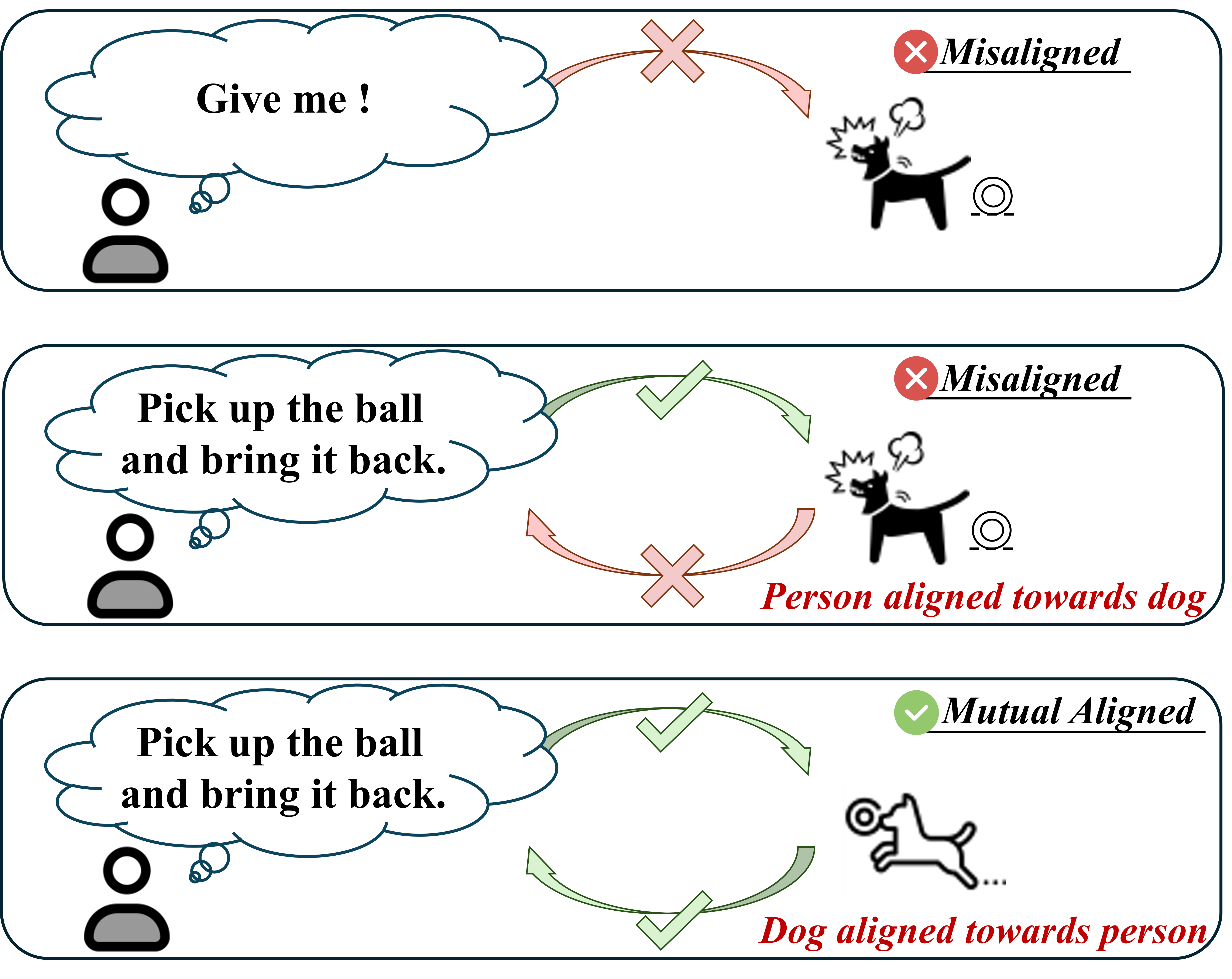}
    \caption{This figure illustrates a common interaction where a person and a dog adjust their behaviors to align instruction with response, evolving through repeated interactions to achieve mutual understanding.}
    \label{fig:instruction_alignment}
\end{figure}

\section{Introduction}
Large Language Models have demonstrated unprecedented capabilities in comprehending human intent and performing cross-task generalization through contextual learning\citep{brown2020language}. A key breakthrough in aligning model behaviors with human expectations is primarily attributed to instruction tuning, a supervised learning paradigm that bridges the gap between pre-trained models' latent knowledge and explicit task requirements \citep{ouyang2022training}. Through multi-task training on (instruction, response) pairs, this approach enables systematic knowledge elicitation while maintaining task-agnostic generalization \citep{chung2024scaling}. The effectiveness of this process is significantly influenced by the availability of high-quality instruction-response pairs at scale. In essence, the quality of data used in instruction tuning is critical to determining the performance and overall effectiveness of the model.

Instruction-tuning methods currently follow two primary approaches. The first involves engaging domain experts \citep{kopf2024openassistant, conover2023free, bach2022promptsource} to manually create instructions for specific tasks, ensuring high precision but facing challenges related to scalability and cost. The second approach \citep{wang2022self, peng2023instruction} leverages LLMs to generate responses based on given prompts. Although this approach is more scalable, it risks introducing inaccuracies or hallucinations \citep{zhang2023language}.

Recent research has explored an alternative: leveraging human-written documents as typical responses and using LLMs to infer user instructions \citep{koksal2023longform, li2023self, chen2024dog, nguyen2024better}, a process known as instruction back-translation. These approaches primarily focused on making the generated data resemble human data, without considering the inherent relationship between the instruction and the response. We contend that the alignment between the instruction and the response is also essential.

As shown in Figure~\ref{fig:instruction_alignment}, the interaction between a person and a dog illustrates the bidirectional nature of training. Both the person and the dog adjust their behaviors to achieve mutual alignment. Similar to how a good command to a dog is one that elicits a proper response, generating an instruction-response pair must be aligned for optimal effectiveness. The quality of the instruction is validated by the response it triggers, and the same logic applies in reverse. Generating a high-quality pair requires careful alignment through mutual interaction. The instruction must clearly guide the response, while the response should accurately reflect the instruction, ensuring that both are mutually reinforcing.

The interdependence between instructions and responses introduces a dual-variable optimization problem, where enhancing one component necessitates adjusting the other simultaneously, as neither can be optimized in isolation. Drawing inspiration from the alternating update strategy used in Expectation-Maximization~(EM) algorithms \citep{moon1996expectation}, we propose \textbf{MAIN}, a framework for synthesizing high-quality data. This framework iteratively optimizes both instructions and responses, progressively reinforcing their mutual alignment. Through this co-adaptive process, the alignment between instruction-response pairs improves substantially, which we believe will significantly boost the model’s performance. Furthermore, we propose a straightforward but effective filtering strategy, \textbf{mutual filter}, which selects pairs with superior alignment, ultimately boosting the quality of the fine-tuning dataset. 

To validate the effectiveness of our proposed MAIN framework, we conducted extensive evaluations by fine-tuning models with instruction-response pairs generated by MAIN across multiple benchmarks. Experimental results demonstrate substantial improvements in output preference, instruction-following capability, and reasoning ability. Specifically, for the LLaMA-2-7B model, our framework achieves a 5.85\% increase in output preference compared to Dog Instruct \citep{chen2024dog}, and a 3.60\% improvement in instruction-following ability over Better Alignment \citep{nguyen2024better}. Furthermore, additional analyses, including experiments on filtering strategies and GPT-4-based pairwise evaluations of instruction alignment, confirm that MAIN’s mutual alignment enhances the coherence and quality of instruction-response pairs.
Our primary contributions are as follows:

\begin{itemize}
    \item We emphasize the critical importance of mutual alignment between instructions and responses in synthesizing high-quality instruction-tuning data.
    \item We propose MAIN, a mutual alignment framework that reinforces the inner connection between instructions and responses, and develop a straightforward but efficient data filtering method.
    \item We conduct extensive evaluations across diverse model families and parameter scales, showing that MAIN outperforms existing methods in enhancing instruction tuning effectiveness. 
\end{itemize}

\section{Methodology}
In this section, we present our proposed Mutual Alignment Framework, designed to enhance instruction tuning performance by establishing and strengthening the intrinsic alignment between instructions and responses.

\begin{figure*}[h] 
    \centering
    \resizebox{0.7\linewidth}{!}{\includegraphics{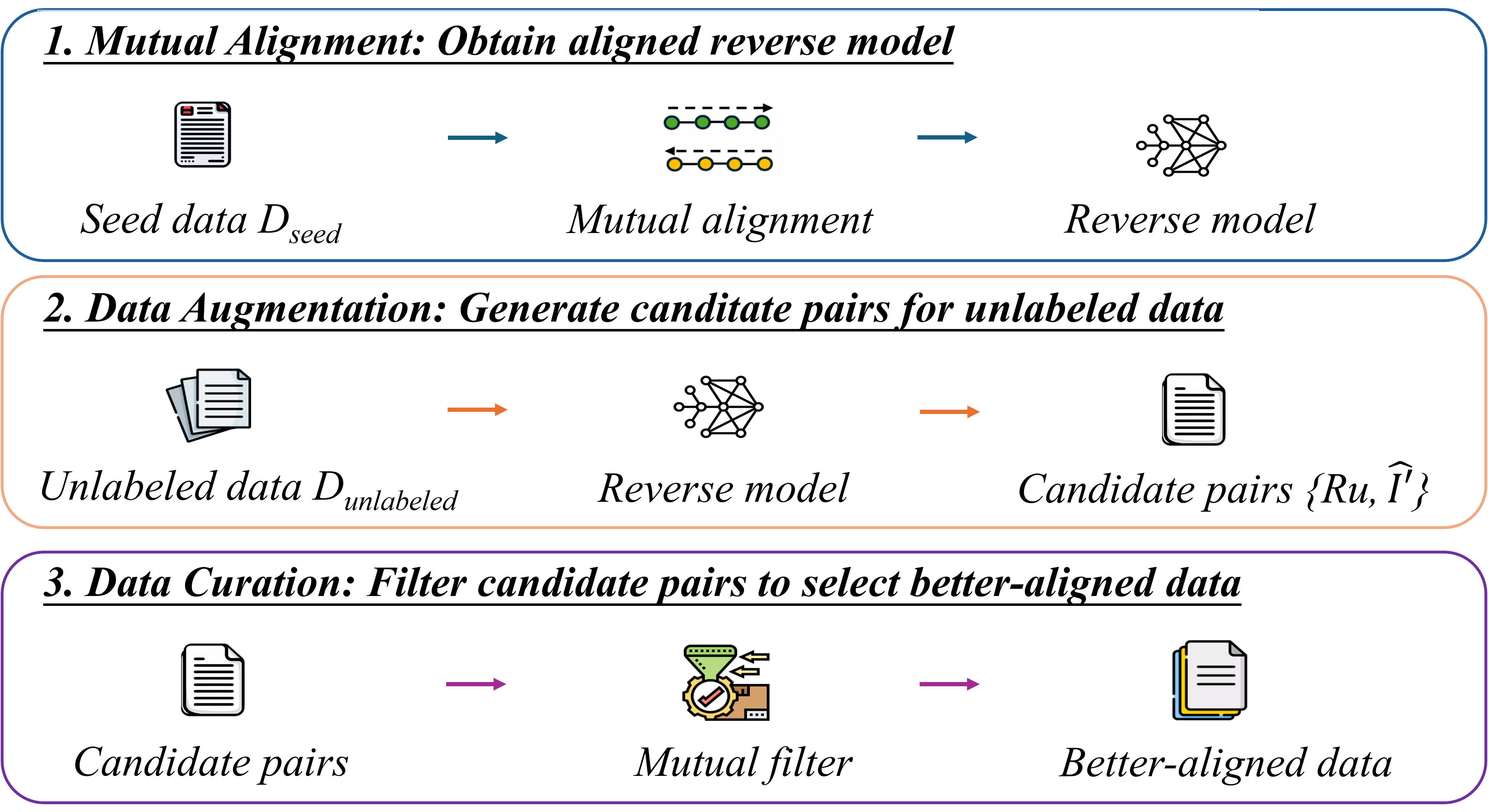}}
    \caption{An overview of the data synthesis process, including mutual alignment, data augmentation, and data curation, aimed at creating high-quality, well-aligned instruction-response pairs from both seed and unlabeled data.}
    \label{fig:overview}
\end{figure*}

\subsection{Preliminary}
\paragraph{Data} 
The framework utilizes two primary datasets: a limited set of high-quality, human-annotated instruction-response pairs seed data \(D_{\text{seed}} = \{(I, R)\}\) and a larger collection of unlabeled responses \(D_{\text{unlabeled}} = \{R_u\}\), extracted from web corpus.
\paragraph{Models} 
The forward model \(M_f := p(R|I)\) is designed to follow instructions, generating responses given instructions, while the reverse model \(M_r := p(I|R)\) learns to generate instructions given responses.

\subsection{Data Synthesis Framework: MAIN}
We present our data synthesis framework, MAIN, illustrated in Figure~\ref{fig:overview}. Given a base language model, a small set of high-quality seed pairs, and a large collection of unlabeled responses, MAIN constructs a high-quality training dataset through three tightly coupled stages: \textbf{Mutual Alignment}, \textbf{Data Augmentation}, and \textbf{Data Curation}.

\begin{itemize}
    \item Mutual Alignment: This step is to obtain a reverse model  \(M_r := p(I|R)\) from the seed data \(D_{\text{seed}}\) based on the base model \(M_{\text{base}}\). This step would align the internal relationship between instruction and response. 
    
    \item Data Augmentation: With the reverse model \(M_r\) trained in the previous step, we apply it to the unlabeled response data \(\mathcal{D}_{\text{unlabeled}}\) to generate corresponding pseudo-instructions. This yields a set of candidate pairs \(\mathcal{D}_{\text{aug}} = \{ (R_u, \hat{I}') \}\), expanding the data space beyond the original seed data.
    
    \item Data Curation: Not all augmented pairs are equally reliable. To select high-quality examples, we apply our mutual filter, which uses both the forward and reverse models to assess alignment consistency. Only examples that meet mutual alignment criteria are retained. These filtered pairs, combined with the original seed data, form the final fine-tuning dataset: \(\mathcal{D}_{\text{filter}} = \text{filter}(\mathcal{D}_{\text{aug}}) \cup \mathcal{D}_{\text{seed}}\).

\end{itemize}

\subsection{Mutual Alignment} 
Achieving strong alignment between instructions and responses is critical for effective instruction tuning. However, establishing a robust relationship between these two components presents a challenging dual-variable problem, as neither direction can be optimized in isolation. Inspired by the iterative principles of the Expectation-Maximization algorithm, we propose mutual alignment that treats instruction-to-response and response-to-instruction generation as complementary tasks, modeled as a forward generation process and a reverse generation process, respectively.
 By alternately optimizing one direction while regulating the other, our method iteratively minimizes discrepancies until convergence is reached, ultimately yielding a model that produces highly aligned instruction–response pairs.

\begin{figure*}[htbp] 
    \centering    \includegraphics[width=1.0\textwidth]{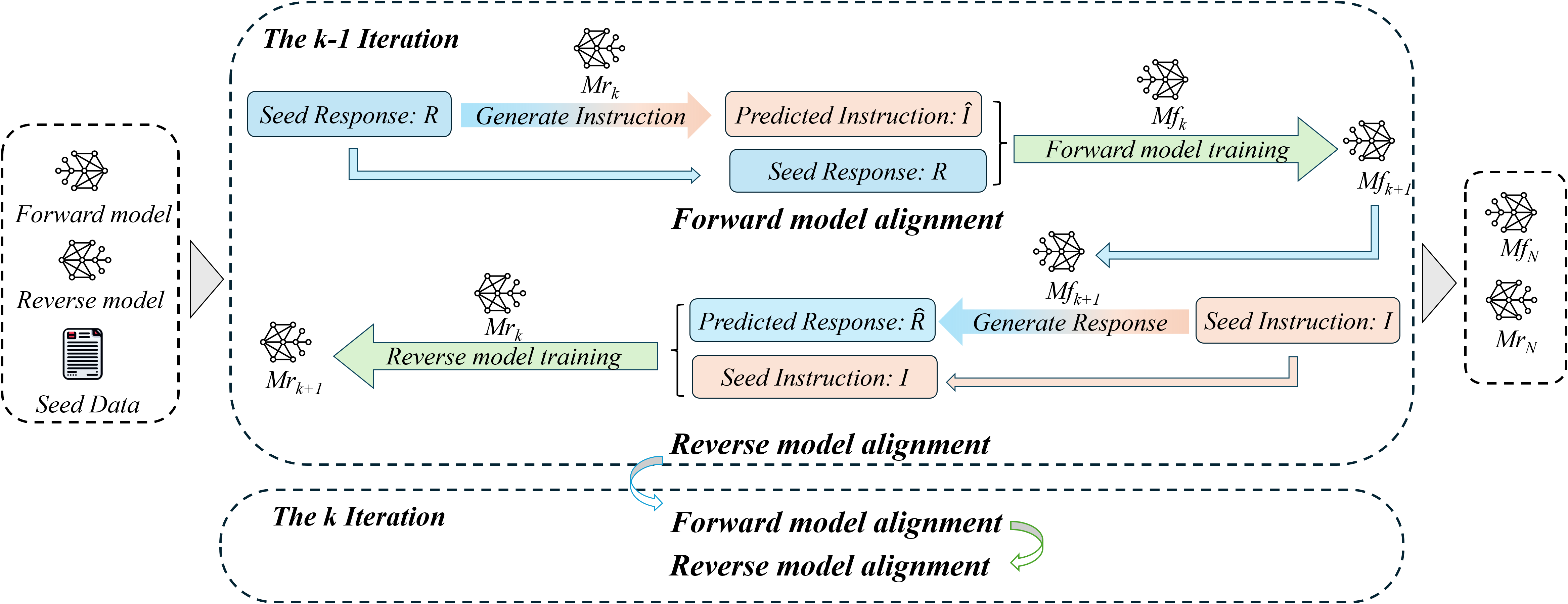} 
    \caption{An overview of our method for iteratively aligning instructions and responses through mutual optimization. }
    \label{fig:example}
\end{figure*}

An overview of our approach is provided in Figure~\ref{fig:example}, and Algorithm~\ref{algorithm:mutual-align} details the iterative optimization process.

\begin{algorithm}[!htb]
    \renewcommand{\algorithmicrequire}{\textbf{Input:}}
    \renewcommand{\algorithmicensure}{\textbf{Output:}}
    \caption{Mutual Alignment}
    \label{algorithm:mutual-align}
    \begin{algorithmic}[1]
        \REQUIRE 
            Seed data $\mathcal{D}_{\text{seed}} = \{(I, R)\}$, Unlabeled data $\mathcal{D}_{\text{unlabeled}} = \{R_u\}$, Base model $M_{\text{base}}$, Number of iterations $N$
        \ENSURE 
            Reverse model $M_r^N$, forward model $M_f^N$
        \STATE Initialize $M_f^0 \leftarrow M_{\text{base}}$, $M_r^0 \leftarrow M_{\text{base}}$
        \FOR{$k = 0$ \TO $N-1$}
            \STATE Generate $\hat{I}$ from $R$ in $\mathcal{D}_{\text{seed}}$ using $M_r^k$
            \STATE Build training set $\mathcal{D}_f = \{(\hat{I}, R)\} \cup \mathcal{D}_{\text{seed}}$
            \STATE Update $M_f^k$ on $\mathcal{D}_f$ by minimizing loss $\mathcal{L}_f$ (Equation~\eqref{eq:2}) to obtain $M_f^{k+1}$
            \STATE Generate pseudo-responses $\hat{R}$ from $I$ in $\mathcal{D}_{\text{seed}}$ using $M_f^{k+1}$
            \STATE Build training set $\mathcal{D}_r = \{(\hat{R}, I)\} \cup \mathcal{D}_{\text{seed}}$
            \STATE Update $M_r^k$ on $\mathcal{D}_r$ by minimizing loss $\mathcal{L}_r$ (Equation~\eqref{eq:4}) to obtain $M_r^{k+1}$
        \ENDFOR
        \STATE \textbf{Return} $M_r^N$ and $M_f^N$
    \end{algorithmic}
\end{algorithm}

\paragraph{Forward Model Alignment.}
To capture the alignment from responses to instructions, we let the forward model learn the distribution that the reverse model simulates. Specifically, at each iteration \(k\), the reverse model generates synthetic instructions \(\hat{I}\) for the responses, forming a target distribution that reflects how instructions should ideally relate to responses. The forward model is then trained to approximate this distribution.


\begin{equation}
\hat{I} = M_r^k(R), \quad \forall R \in \mathcal{D}_{\text{seed}}.
\end{equation}
These synthetic pairs \((\hat{I}, R)\) are merged with the original seed data \((I, R)\) to form the training set. The forward model is then updated to \(M_f^{k+1}\) by optimizing a weighted loss function:

\begin{equation}
\mathcal{L}_f = \alpha \cdot \mathcal{L}(\hat{I}, R) + (1-\alpha) \cdot \mathcal{L}(I, R).
\label{eq:2}
\end{equation}
The first loss term \(\mathcal{L}(\hat{I}, R)\) aligns the forward model with the synthetic instructions generated by the reverse model, ensuring that the forward model learns how responses correspond to instructions as modeled by the reverse model. The second loss term \(\mathcal{L}(I, R)\) maintains consistency with the original human-annotated instructions, thereby preventing the forward model from overfitting to synthetic data. The parameter \(\alpha\) controls the balance between synthetic and human-annotated instructions, with its dynamic adjustment described in Dynamic Weighting. This process encourages the forward model to adapt to the instruction distribution induced by the reverse model.

\paragraph{Reverse Model Alignment.} 
Similarly, the reverse model is trained to capture the alignment from instruction to response as guided by the forward model. 
The reverse model now is updated based on the latest forward model \(M_f^{k+1}\) that generates synthetic responses \(\hat{R}\) conditioned on the seed instructions:

\begin{equation}
\hat{R} = M_f^{k+1}(I), \quad \forall I \in \mathcal{D}_{\text{seed}}.
\end{equation}

And it is optimized using similar weighted loss function:

\begin{equation}
\mathcal{L}_r = \alpha \cdot \mathcal{L}(\hat{R}, I) + (1-\alpha) \cdot \mathcal{L}(R, I).
\label{eq:4}
\end{equation}

\paragraph{Dynamic Weighting}
\label{dynamic}
To balance the influence of synthetic and seed data, we adopt a dynamic weighting strategy that adaptively adjusts their contributions during training. The weighting coefficient \(\alpha \in [0, 1]\) controls this balance, where static settings may lead to suboptimal outcomes: over-reliance on synthetic data can introduce noise, while overemphasis on seed data may hinder generalization. To address this, we update \(\alpha\) at each step based on the relative loss of the two data sources. For forward model training, the update rule is:

\begin{equation}
\alpha = \frac{\mathcal{L}(\hat{I}, R)}{\mathcal{L}(\hat{I}, R) + \mathcal{L}(I, R)}.
\label{eq:5}
\end{equation}

This formulation ensures that the contribution of synthetic data is modulated according to its training loss, enabling the model to incorporate novel patterns from synthetic pairs while preserving the stability offered by seed data.

\subsection{Data Augmentation}
Following mutual alignment optimization, we expand the training corpus by generating synthetic instructions for unlabeled responses. Specifically, for each unlabeled response \(R_u \in \mathcal{D}_{\text{unlabeled}}\), the reverse model generates a corresponding synthetic instruction \(\hat{I}'\), forming a set of candidate instruction–response pairs \(\{(R_u, \hat{I}')\}\). These pairs approximate how users might naturally formulate instructions for the given responses. However, as the quality of generated pairs may vary, a subsequent curation step is required to ensure alignment consistency and data reliability.

\subsection{Data Curation}
To further improve data alignment, we introduce an effective filtering mechanism. We assume that high-quality instruction-response pairs should be well-aligned, where the predicted instruction generated by the reverse model can be decoded by the forward model to recover the response, which should closely resemble the original. This process is akin to the interaction between an encoder and a decoder \citep{cho2014learning}. Thus, we select the most well-aligned pairs. Using the candidate pairs \(\{R_u, \hat{I}'\}\) from the Augmentation stage, we then employ the forward model to generate synthetic responses \(\hat{R}'\) based on \(\hat{I}'\):
\begin{equation}
\hat{R}' = M_f^N(\hat{I}').
\end{equation}
We compute the Cross-Entropy between the synthetic responses \(\hat{R}'\) and the original unlabeled responses \(R_u\):
\begin{equation}
\mathcal{L}_{\text{CE}}(\hat{R}', R_u) = - \sum_{i} \log p(\hat{R}' \mid \hat{I}', R_u).
\end{equation}
Candidate pairs are sorted in ascending order by their values, and only those with the smallest values—indicating the highest degree of mutual alignment are retained.:
\begin{equation}
\mathcal{D}_{\text{filter}} = \left[ \text{filter}(\mathcal{D}_{\text{aug}}), \mathcal{D}_{\text{seed}} \right]
\end{equation}
This straightforward mechanism, relying solely on our mutual alignment model, effectively curates a high-quality subset of data for fine-tuning.


\section{Experiment}
\subsection{Experimental Setup}

\paragraph{Data.}
The seed data consists of 3,200 human-annotated (instruction, response) examples from the Open Assistant dataset \citep{kopf2024openassistant}, serving as a reliable baseline for fine-tuning. The unlabeled data is Falcon RefinedWeb \citep{penedo2023refinedweb} that is a massive English web dataset containing raw responses without paired instructions. We sampled 502k segments.

\paragraph{Mutual Alignment Framework.}
For mutual alignment experiments, we adopt LLaMA-2-7B as the base model, and additionally evaluate the generalization of our approach on Mistral and Qwen models. In each iteration, both the forward and reverse models are trained for one epoch. We use a learning rate of $1 \times 10^{-5}$ with a linear decay schedule. The adaptation weight \(\alpha\) is dynamically updated according to Equation~\ref{eq:5}. The batch size is set to 32. For data curation, we select the top 16,800 instruction-response pairs from the unlabeled set based on cross-entropy scores and combine them with the seed data to form the final fine-tuning dataset.

\paragraph{Base model \& fine-tuning.}  
We use the pre-trained LLaMA, Mistral and Qwen as the base models respectively. Detailed hyperparameter configurations are provided in Appendix~\ref{appendix:training_details}.



\subsection{Benchmarks}
To evaluate our framework, we conduct experiments across three benchmarks that assess different aspects of model performance.
\paragraph{AlpacaEval.}  
We assess output preference using 805 instructions from the AlpacaEval dataset~\citep{li2023alpacaeval}. Model outputs are compared against text-davinci-003 in a pairwise setting, with GPT-4-based judgments determining win rates.

\paragraph{IFEval.}  
Instruction-following ability is evaluated with IFEval~\citep{zhou2023instruction}, which reports accuracy across four metrics: Prompt-level Strict (P-S), Instruction-level Strict (I-S), Prompt-level Loose (P-L), Instruction-level Loose (I-L) ensuring a comprehensive assessment of instruction adherence.

\paragraph{OpenLLM.}  
Reasoning ability is measured via the Open LLM Leaderboard~\citep{open-llm-leaderboard-v1} using the Language Model Evaluation Harness~\citep{eval-harness}. We evaluate on ARC~\citep{clark2018think}, HellaSwag~\citep{zellers2019hellaswag}, Winogrande~\citep{sakaguchi2021winogrande}, MMLU~\citep{hendrycks2020measuring}, and TruthfulQA~\citep{lin2021truthfulqa}.

\subsection{Baselines}

We compare our framework to several baseline approaches, fine-tuned on 3.2k seed data and 16.8k generated data from Falcon-RefinedWeb dataset.

\paragraph{Longform.}  
This method \citep{koksal2023longform} prompts a large language model to generate instructions for human-written texts.

\paragraph{Humpback.}  
This method \citep{li2023self} is a two-stage curation process that filters and selects high-quality pairs before fine-tuning.

\paragraph{Dog Instruct.}  
This method \citep{chen2024dog} involves a post-processing step that refines the responses to align with standard AI-generated output.

\paragraph{Better Alignment.}  
This method \citep{nguyen2024better} generates instructions via back-translation, then filters pairs to obtain high-quality response.

\begin{table*}[htbp]
\centering
\resizebox{\textwidth}{!}{
    \begin{tabular}{lccccccccccc}
    \toprule
    \multicolumn{1}{c}{\textbf{Base Model}} & \multicolumn{1}{c}{\textbf{Method}} & \multicolumn{1}{c}{\textbf{Output preference}} & \multicolumn{4}{c}{\textbf{Instruction following}} & \multicolumn{5}{c}{\textbf{Reasoning ability}} \\
    \cmidrule(r){1-1} \cmidrule(r){2-2} \cmidrule(r){3-3} \cmidrule(r){4-7} \cmidrule(r){8-12} 
    & & \multirow{2}{*}{\makecell{AlpacaEval}} & \multicolumn{4}{c}{IFEval} & \multirow{2}{*}{\makecell{ARC\_C}} & \multirow{2}{*}{\makecell{MMLU}} & \multirow{2}{*}{\makecell{HellaSwag}} & \multirow{2}{*}{\makecell{Winogrande}} & \multirow{2}{*}{\makecell{TruthfulQA}} \\ \cmidrule(r){4-7}
    & & & \makecell{P-S} & \makecell{I-S} & \makecell{P-L} & \makecell{I-L} & & & & & \\
\midrule
\multirow{6}{*}{Llama-2-7B} 
    & Humpback \citep{li2023self} & 41.02 & 15.46 & 26.42 & 18.39  & 29.91 & 55.90 & 44.91 & 79.42 & 73.32 & 44.48   \\
    & Longform \citep{koksal2023longform} & 35.64 & 15.23 & 26.10 & 17.56  & 29.29 & 55.72 & 45.02 & 78.98 & 73.21 & 45.07  \\
    & Dog Instruct\citep{chen2024dog} & \underline{52.35} & 15.52 & \underline{28.17} & 19.40 &   \underline{32.01} & \underline{56.06} & 45.62 & 79.89 & \underline{74.13} & \underline{45.77} \\
    & Better Alignment \citep{nguyen2024better} & 50.37 & \underline{16.82} & 27.70 & \underline{19.69}  & 31.52 & 55.92 & \underline{\textbf{45.84}} & \underline{80.33} & 74.12 & 45.30  \\
    & \textbf{MAIN} & \textbf{58.20} & \textbf{20.22}  & \textbf{31.17} & \textbf{23.36}    & \textbf{35.37}  & \textbf{57.08} & 45.47 & \textbf{81.22} & \textbf{74.51} &\textbf{ 47.40} \\
    \rowcolor{blue!10} 
    & $\Delta$ over Best Result &  +5.85 & +3.40 & +3.00 & +3.67  & +3.36 & +1.02 & -0.37 & +0.89 & +0.38 & +1.63 \\
\midrule
\multirow{6}{*}{Mistral-7B} 
    & Humpback \citep{li2023self} & 40.48 & 17.19 & 28.05 & 20.88 & 32.37 & 54.01 & 49.26 & 79.12 & \underline{\textbf{73.24}} & 45.48  \\
    & Longform \citep{koksal2023longform} & 37.62 & 16.98 & 27.89 & 20.75 & 32.10 & 53.98 & 48.12 & 78.25 & 71.60 & 44.88  \\
    & Dog Instruct\citep{chen2024dog} & 45.34 & 18.23 & 28.48 & 21.32 & 33.47 & 53.15 & 49.10 & \underline{79.07} & 73.21  & \underline{45.98} \\
    & Better Alignment\citep{nguyen2024better} & \underline{45.79} & \underline{18.35} & \underline{29.45} & \underline{21.49} & \underline{34.10} & \underline{54.20} & \underline{50.28} & 78.37 & 71.48 & 44.73  \\
    & \textbf{MAIN} & \textbf{48.94} & \textbf{23.47} & \textbf{34.29} &\textbf{ 26.60} & \textbf{38.84} & \textbf{55.12} &\textbf{ 52.93} & \textbf{79.38} & 72.38 & \textbf{49.76} \\
    \rowcolor{blue!10} 
    & $\Delta$ over Best Result &  +3.15 & +5.12 & +4.84 & +5.11 & +4.74 & +0.92 & +2.65 & +0.31 & -0.86 & +3.78 \\
\midrule
\multirow{6}{*}{Qwen2.5-14B}
    & Humpback \citep{li2023self}                   & 49.83& 69.45 & 79.92 & 74.28  & 82.25      & 65.25 & 77.04 & 83.81 & 78.91 & 57.63 \\
    & Longform \citep{koksal2023longform}           & 45.42 & 69.12 & 79.37 & 74.05  & 82.46     & 65.12 & 76.92 & 83.04 & 76.55 & 55.72 \\
    & Dog Instruct \citep{chen2024dog}              & 50.27 & \underline{72.83} & \underline{81.89} & \underline{76.64}  & \underline{84.72}     & 64.71 & \underline{78.33} & 84.19 & 79.12 & \underline{56.34} \\
    & Better Alignment \citep{nguyen2024better}     & \underline{53.93} & 71.37 & 81.55 & 76.32  & 84.14     & \underline{65.36} & 78.01 & \underline{\textbf{84.25}} & \underline{79.63} & 55.91 \\
    & \textbf{MAIN}                             & \textbf{61.30} & \textbf{74.90} & \textbf{84.03} & \textbf{79.93}  & \textbf{87.61}     & \textbf{66.93}  & \textbf{80.26} & 84.07 & \textbf{80.72} & \textbf{59.49} \\
    \rowcolor{blue!10} 
    & $\Delta$ over Best Result & +7.37 & +2.07 & +2.14 & +2.29  & +2.89  & +1.57 & +1.93 &  -0.18 & +1.09 & +3.15 \\

\bottomrule
\end{tabular}}  
\caption{Benchmarking results of different methods on Llama-2-7B, Mistral-7B and Qwen2.5-14B using Falcon RefinedWeb dataset given same data quantity (20k samples). $\Delta$ over Best Result quantify improvements relative to the strongest baseline method across evaluation categories.}
\label{tab:llama_mistral_comparison}
\end{table*}

\section{Experimental Results}
This section presents the experimental results, including quantitative results, an ablation study, data alignment analysis and a case study, to assess the effectiveness of our approach.
\subsection{Quantitative Results}
We conduct experiments across three benchmarks, each assessing a different aspect of the MAIN: AlpacaEval evaluates output preference, IFEval measures instruction-following ability, and OpenLLM tests reasoning capability, with additional experiments detailed in Appendix~\ref{appendix:Generalization_Results}. For completeness, we also provide additional results on multilingual generalization in Appendix~\ref{appendix:Multilingual_Generalization}.
\paragraph{Output Preference.}
As shown in Table~\ref{tab:llama_mistral_comparison}, our method achieves the highest win rate in AlpacaEval dataset, surpassing the leading baseline. Specifically, on Llama-2-7B, our method achieves a win rate of 58.20\%, representing a 5.85\% improvement over the best baseline Dog Instruct \citep{chen2024dog}. On Mistral-7B, our method outperforms Better alignment \citep{nguyen2024better} by 3.15\%, reaching a win rate of 48.94\% compared to 45.79\%. On Qwen2.5-14B, MAIN achieves the highest win rate of 71.30\%. These results confirm that our MAIN method enhances instruction-response alignment more effectively than previous approaches, leading to outputs that better align with human expectations.

\paragraph{Instruction Following.}
Table~\ref{tab:llama_mistral_comparison} illustrates the results of our method in IFEval dataset where MAIN achieves state-of-the-art performance across all three model backbones. Compared to the best-performing baseline, our approach achieves consistent improvements across all evaluation metrics. Specifically, on Llama-2-7B, we see an increase of 2.59\% in P-S and 3.42\% in I-S. For Mistral-7B, we observe a 5.12\% improvement in P-S and a 4.84\% improvement in I-S over Better alignment \citep{nguyen2024better}. For Qwen2.5-14B, MAIN also leads with a +2.07\% and +2.99\% gain in P-S and I-S respectively. These results highlight the crucial role of enhanced data alignment in fine-tuning, which allows our model to better interpret and respond to user instructions, thereby driving its superior performance in instruction-following tasks.

\paragraph{Reasoning Ability.}
As shown in Table~\ref{tab:llama_mistral_comparison}, our method demonstrates strong improvements in reasoning and factual accuracy across multiple downstream tasks. On Llama-2-7B, our approach shows a 2.02\% improvement over the best baseline, Better Alignment, on ARC-Challenge and a 1.63\% improvement on TruthfulQA. In particular, ARC-Challenge benefits from our method's ability to better capture common-sense reasoning patterns, which likely leads to more accurate responses. On Mistral-7B, the most significant improvements are observed in TruthfulQA, where our method outperforms Better Alignment by 5.03\%, and in MMLU, with a 2.65\% increase. We further include evaluation on Qwen2.5-14B, which follows a similar trend. These benchmarks that require accurate factual recall and complex reasoning, show how our method strengthens the model's ability to provide correct and contextually appropriate answers.


\begin{figure*}[t]
    \centering
    \label{fig:openllm2}
\end{figure*}

\subsection{Ablation Study}

We conduct additional ablation study to analyze the impact of our filtering strategy. Further ablation experiments are provided in Appendix~\ref{appendix:additional_ablations}.

\paragraph{Filtering Stragety.}

Effective filtering is critical for improving alignment quality by removing noisy or misaligned instruction–response pairs. Table~\ref{tab:filtering} compares models trained with no filtering, score-based filtering, and our proposed mutual-filter method. Among these, our mutual-filter achieves the highest win rate and delivers the most consistent gains in instruction-following accuracy.

In contrast, score-based filtering provides little benefit and even underperforms compared to unfiltered data. This is due to the score-based approach, used in Humpback \citep{li2023self} and Better Alignment \citep{nguyen2024better}, relying on a ranking model fine-tuned on seed data rather than a dedicated scoring model. Without a clear optimization objective for instruction-response alignment, it struggles to identify high-quality pairs, leading to suboptimal fine-tuning.

Our mutual filter, by leveraging mutual-alignment models, directly favors instruction-response pairs with strong coherence. By eliminating misaligned samples without requiring additional supervision, it ensures a more effective training dataset, resulting in improved instruction-following and generalization capabilities.

\begin{table}[htbp]
\centering
\resizebox{\columnwidth}{!}{
\renewcommand{\arraystretch}{1.2} 
\begin{tabular}{lccccc}
\toprule
\textbf{Filtering Method} & \textbf{Win Rate} & \textbf{P-S} & \textbf{I-S} & \textbf{P-L} & \textbf{I-L} \\
\midrule
Ours w/o filtering & 56.40 & 19.41 & 23.11 & 29.74 & 34.17 \\
Ours w/ score-based filtering & 55.26 & 17.63 & 20.21 & 29.11 & 33.72 \\
Ours w/ mutual filter & 58.20 & 20.22 & 23.36 & 31.17 & 35.37 \\
\bottomrule
\end{tabular}
}
\caption{Performance evaluation of LLaMA-2-7B fine-tuned on the Falcon-RefinedWeb dataset (20K samples) under three filtering conditions. All conditions operate on instruction-response pairs generated by the same reverse model. }
\label{tab:filtering}
\end{table}

\begin{figure*}[htbp]
    \centering
    \includegraphics[width=0.90\linewidth]{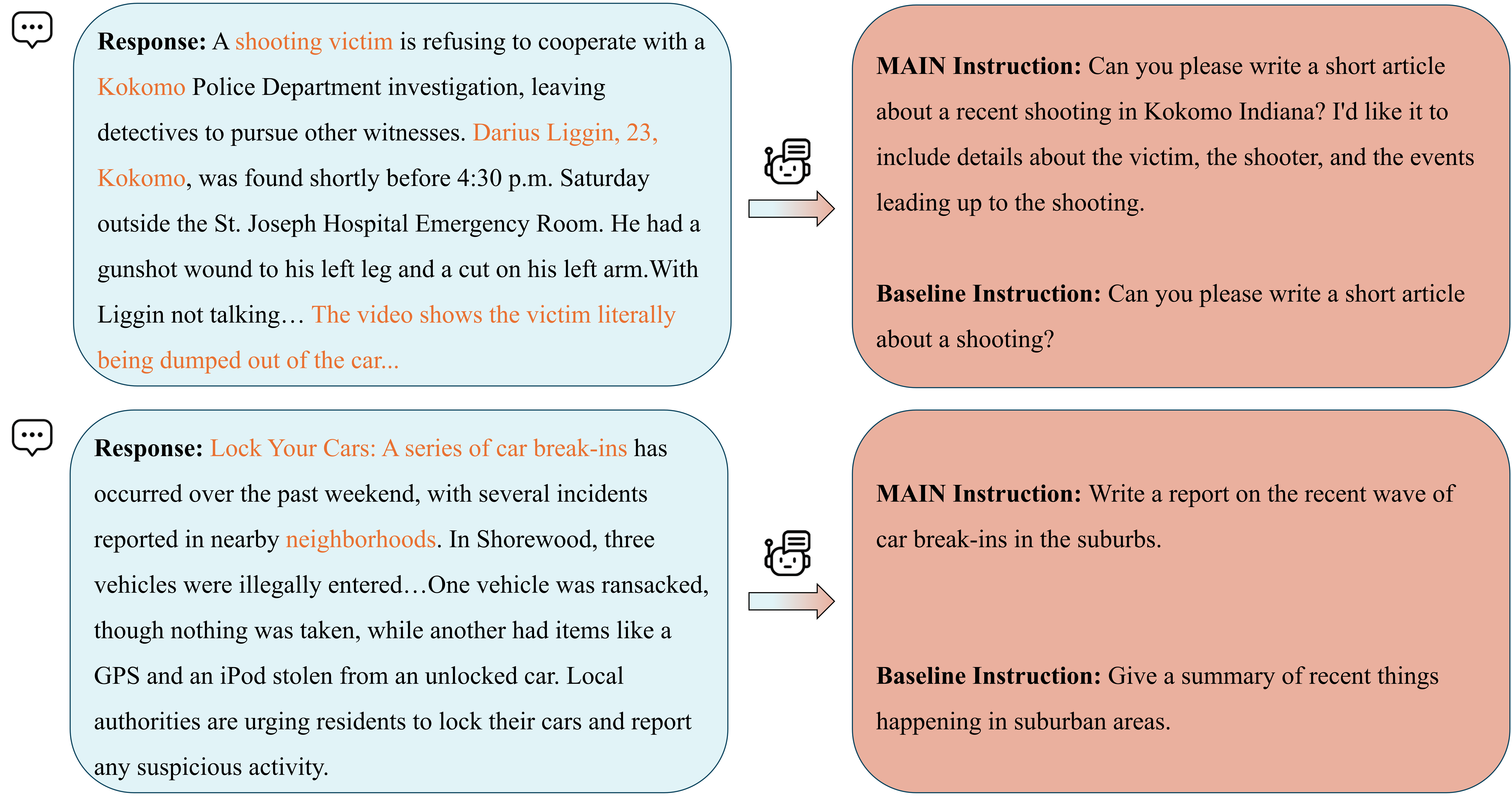} 
    \caption{Method Comparison for Instruction Generation: A Case Study on the Effectiveness of Reverse Model Approaches in Aligning Instructions with Responses}
    \label{fig:Case study}
\end{figure*}

\subsection{Data Alignment Analysis}
We assessed the alignment quality of instructions generated by our MAIN method compared to baselines using blind pairwise evaluations conducted by GPT-4. Specifically, we randomly selected 1000 responses from the Falcon RefinedWeb dataset. For each response, GPT-4 evaluated two candidate instructions—one from MAIN and one from a baseline—in random order to avoid positional bias. We then calculated Win, Tie, and Loss rates based on GPT-4’s judgments: Win indicates GPT-4 preferred MAIN, Tie indicates no clear preference, and Loss indicates GPT-4 preferred the baseline. Using the Qwen2.5-14B as the base model, MAIN consistently outperformed all baselines, with win rates ranging from 61.7\% to 81.6\%, demonstrating superior instruction alignment capability (see Table~\ref{tab:alignment_quality}). The evaluation prompt used by GPT-4 is detailed in Appendix~\ref{appendix:gpt4-prompt}.

\begin{table}[htbp]
\centering
\renewcommand{\arraystretch}{1.2}
\resizebox{\columnwidth}{!}{
\begin{tabular}{lccccc}
\toprule
\textbf{Baseline} & \textbf{Win Rate} & \textbf{Tie Rate} & \textbf{Loss Rate} & \textbf{$\Delta$} \\
\midrule
Humpback         & 69.3\% & 18.6\% & 12.1\%  & +56.2\% \\
Longform         & 81.6\% & 8.6\%  & 8.8\%   & +72.8\% \\
Dog Instruct     & 61.7\% & 15.1\% & 23.2\%  & +38.5\% \\
Better Alignment & 64.7\% & 12.4\% & 22.9\%  & +41.8\% \\
\bottomrule
\end{tabular}
}
\caption{
GPT-4 pairwise evaluation comparing \textbf{MAIN} with four baseline methods on instruction alignment quality. Each comparison is based on 1000 examples sampled from the Falcon RefinedWeb dataset. $\Delta$ denotes the win rate margin of MAIN over each baseline.
}
\label{tab:alignment_quality}
\end{table}

\subsection{Case Study}
As shown in Figure \ref{fig:Case study}, the two examples, both extracted from unlabeled data, illustrate the effectiveness of our approach. In the first case, the MAIN instruction explicitly requests specific details about the victim, the shooter, and the events surrounding the shooting, providing clear guidance for the response. In contrast, the baseline instruction is more general, asking for a brief article about the shooting without specifying key details. In the second case, the MAIN instruction emphasizes a critical event: a wave of car burglaries in the suburbs, while the baseline instruction remains vague, simply requesting a summary of events in suburban areas.

In both cases, MAIN Instructions are more focused and specific, resulting in responses that are better aligned with the intended context. In contrast, the baseline instructions are more general. These examples demonstrate that our method generates instructions that are more closely aligned with the responses.

\section{Related Work}
\subsection{Instruction Tuning}
Instruction tuning fine-tunes pre-trained LLMs on instruction-response pairs, enabling models to generalize across tasks without task-specific fine-tuning \citep{wei2021finetuned, mishra2021cross, wang2022super}. Subsequent work \citep{mishra2021cross, sanh2021multitask} focused on cross-task generalization through diverse inputs.

\subsection{Data Generation}
Effective instruction tuning relies on large-scale, high-quality datasets, typically generated in two ways: human-crafted or model-generated.

\paragraph{Human-Crafted Data}
Datasets curated by domain experts, like OpenAssistant Conversations \citep{kopf2024openassistant} and Databricks Dolly-15k \citep{conover2023free}, are high quality but costly. Crowd-sourced platforms like ShareGPT \citep{chiang2023vicuna} also contribute valuable data, especially user-uploaded conversations.

\paragraph{Model-Generated Data}
To reduce manual annotation costs, methods like Self-Instruct \citep{wang2022self} and Alpaca-GPT4 \citep{peng2023instruction} generate instruction-response pairs automatically. However, issues like hallucinations \citep{zhang2023language} persist. New approaches, such as Better Alignment \citep{nguyen2024better} and Dog-Instruct\citep{chen2024dog}, pair human responses with inferred instructions to reduce hallucinations and improve scalability. Our proposed MAIN builds on this by iteratively optimizing instruction-response alignment to ensure high-quality data.

\section{Conclusion} 
In this paper, we highlight the critical role of instruction-response alignment in instruction tuning for LLMs. We introduce the Mutual Alignment Framework, which iteratively optimizes both instructions and responses to improve their coherence, along with a mutual filtering strategy to select high-quality pairs. Experiments across multiple benchmarks show that our framework enables state-of-the-art performance. These findings highlight the importance of mutual alignment in instruction tuning and offer a new perspective for refining instruction-response pairs, paving the way for more effective and principled instruction tuning in future LLM development.

\section*{Limitations}
Our experiments cover a broad range of model families and parameter scales, demonstrating the robustness of the proposed framework across architectures and sizes. However, we have not yet evaluated MAIN on very large models due to resource limit, which may exhibit qualitatively different behaviors. This limits our ability to fully assess the potential of mutual alignment at extreme scales. Investigating its effectiveness in such settings remains an important direction for future work.

\bibliography{acl_latex}

\newpage
\appendix
\section{Training Details}

\label{appendix:training_details}
\begin{table}[ht]
\centering
\resizebox{\columnwidth}{!}{
\begin{tabular}{ll}
\hline
\textbf{Hyperparameter} & \textbf{Assignment} \\ \hline
Computing Infrastructure & 8 A100-80GB GPUs \\
Number of epochs & 2 \\
Batch size per GPU & 64 \\
Maximum sequence length & 1024 \\
Maximum learning rate &  2e-5  \\
Optimizer & Adam \\
Adam epsilon & 1e-8 \\
Adam beta weights & 0.9, 0.999 \\
Learning rate scheduler & warmup linear \\
Weight decay & 0.1 \\
Warmup steps & 100 \\
Learning rate decay & linear \\
\hline
\end{tabular}}
\caption{Hyperparameters used in the experiments.}
\label{tab:hyperparameters}
\end{table}
Training is conducted with hyperparameters aligned to established supervised fine-tuning (SFT) practices \citep{zhou2024lima, touvron2023llama}. The learning rate is set to \(2 \times 10^{-5}\), with a weight decay of 0.1, a batch size of 64, and a dropout rate of 0.1. Additionally, each iterative phase of training is limited to one epoch. For text generation, we apply nucleus sampling \citep{holtzman2019curious} with a temperature (\(T\)) of 0.7 and a top-\(p\) value of 0.9. These settings balance diversity and relevance in the generated outputs. More hyperparameters listed in Table \ref{tab:hyperparameters}

\section{Generalization Results} 
\label{appendix:Generalization_Results}
Table~\ref{tab:appendix_generalization_results} presents extended results that demonstrate both the scalability across varying model sizes across diverse architectures of our MAIN method. These capabilities are evidenced by MAIN's enhanced performance compared to the baseline \citep{nguyen2024better} on the IFEval benchmark when evaluated on models such as Qwen2.5-3B and LLaMA-2-13B.

\begin{table}[htbp]
  \centering
  \resizebox{\columnwidth}{!}{%

    \begin{tabular}{@{}ll SSSS@{}}
      \toprule
      \textbf{Model} & \textbf{Method} & {\textbf{P-S (\%)}} & {\textbf{I-S (\%)}} & {\textbf{P-L (\%)}} & {\textbf{I-L (\%)}} \\
      \midrule
      \multirow{2}{*}{Qwen2.5-3B} & Baseline & 32.13 & 35.98 & 44.02 & 47.10 \\
                                 & MAIN   & \bfseries 35.78 & \bfseries 39.12 & \bfseries 47.42 & \bfseries 51.21 \\
      \addlinespace
      \multirow{2}{*}{LLaMA-2-13B} & Baseline & 16.12 & 18.72 & 28.54 & 30.20 \\
                                  & MAIN   & \bfseries 18.98 & \bfseries 22.63 & \bfseries 30.87 & \bfseries 34.85 \\
      
      \bottomrule
    \end{tabular}%
  } 
  \caption{Comparative instruction-following performance of MAIN and Baseline on the IFEval dataset across diverse models.}
  \label{tab:appendix_generalization_results}
\end{table}

Further evaluations in Table~\ref{tab:appendix_diverse_results} demonstrate that our MAIN framework consistently outperforms Baseline \citep{nguyen2024better} across diverse NLP benchmarks (BLEU, ROUGE-L, SQuADv2).

\begin{table}[htbp]
  \centering
  \resizebox{\columnwidth}{!}{%
    \sisetup{table-format=2.2, tight-spacing=true}
    \begin{tabular}{@{}ll S S l@{}}
      \toprule
      \textbf{Base Model} & \textbf{Method} & {\textbf{BLEU}} & {\textbf{ROUGE-L}} & \textbf{SQuADv2 (EM / F1)} \\
      \midrule
      \multirow{2}{*}{Llama-2-7B} & Baseline & 43.94 & 43.82 & 10.82 / 19.62 \\
                                 & MAIN   & \bfseries 47.37 & \bfseries 46.39 & \bfseries 12.99 / 21.30 \\
      \addlinespace
      \multirow{2}{*}{Mistral-7B} & Baseline & 40.75 & 39.41 & 14.53 / 21.72 \\
                                 & MAIN   & \bfseries 43.15 & \bfseries 44.70 & \bfseries 17.11 / 23.49 \\
      \bottomrule
    \end{tabular}%
  }
  \caption{Comparative performance of MAIN and Baseline on diverse NLP benchmarks.} 
  \label{tab:appendix_diverse_results} 
\end{table}

\section{Ablations}  
\label{appendix:additional_ablations}
\paragraph{Training iterations.} 
To analyze the impact of iteration count \(N\), we vary \(N\) from 1 to 20 and evaluate its effect on  AlpacaEval and IFEval dataset. As shown in Table \ref{tab:iteration_ablation}, increasing \(N\) initially improves performance, as iterative refinement enables the forward and reverse models to progressively align their outputs, enhancing instruction-response consistency.  

However, beyond a certain point, performance begins to decline. Excessive iterations reinforce suboptimal patterns leading to overfitting. This underscores the necessity of selecting an optimal \(N\) that balances refinement and generalization. Our results emphasize the importance of properly tuning \(N\) to maximize the benefits of mutual alignment.

\begin{table}[htbp]
\centering
\setlength{\tabcolsep}{10pt} 
\begin{tabular}{lc}
\toprule
\textbf{Iterations \(N\)} & \textbf{Win Rate} \\
\midrule
\(N=1\) & 50.11 \\
\(N=2\) & 55.72 \\
\(N=3\) & 58.20 \\
\(N=4\) & 55.89 \\
\(N=5\) & 55.60 \\
\(N=10\) & 54.41 \\
\(N=15\) & 54.29\\
\(N=20\) & 54.52 \\
\bottomrule
\end{tabular}
\caption{Ablation study on the effect of iteration count \(N\). We analyze the influence of varying the number of training iterations (\(N = 1, 2, 3, 4, 5, 10, 15, 20\)) on Llama-2-7B fine-tuned on Falcon-RefinedWeb.}
\label{tab:iteration_ablation}
\end{table}

\paragraph{Dynamic Weighting.} 
Balancing the contribution of aligned instruction-response pairs is crucial for achieving both strong alignment and robust generalization. The weighting parameter \(\alpha\) controls this balance during training by adjusting the relative influence of synthetic and seed data. To evaluate its effectiveness, we compare fixed values (\(\alpha = 0.3, 0.5, 0.7, 0.8, 1.0\)) with our adaptive approach (\(\alpha=\) Dynamic), which continuously updates \(\alpha\) throughout training.

\begin{figure}[htbp]
\small
    \centering
    \includegraphics[width=0.70\linewidth]{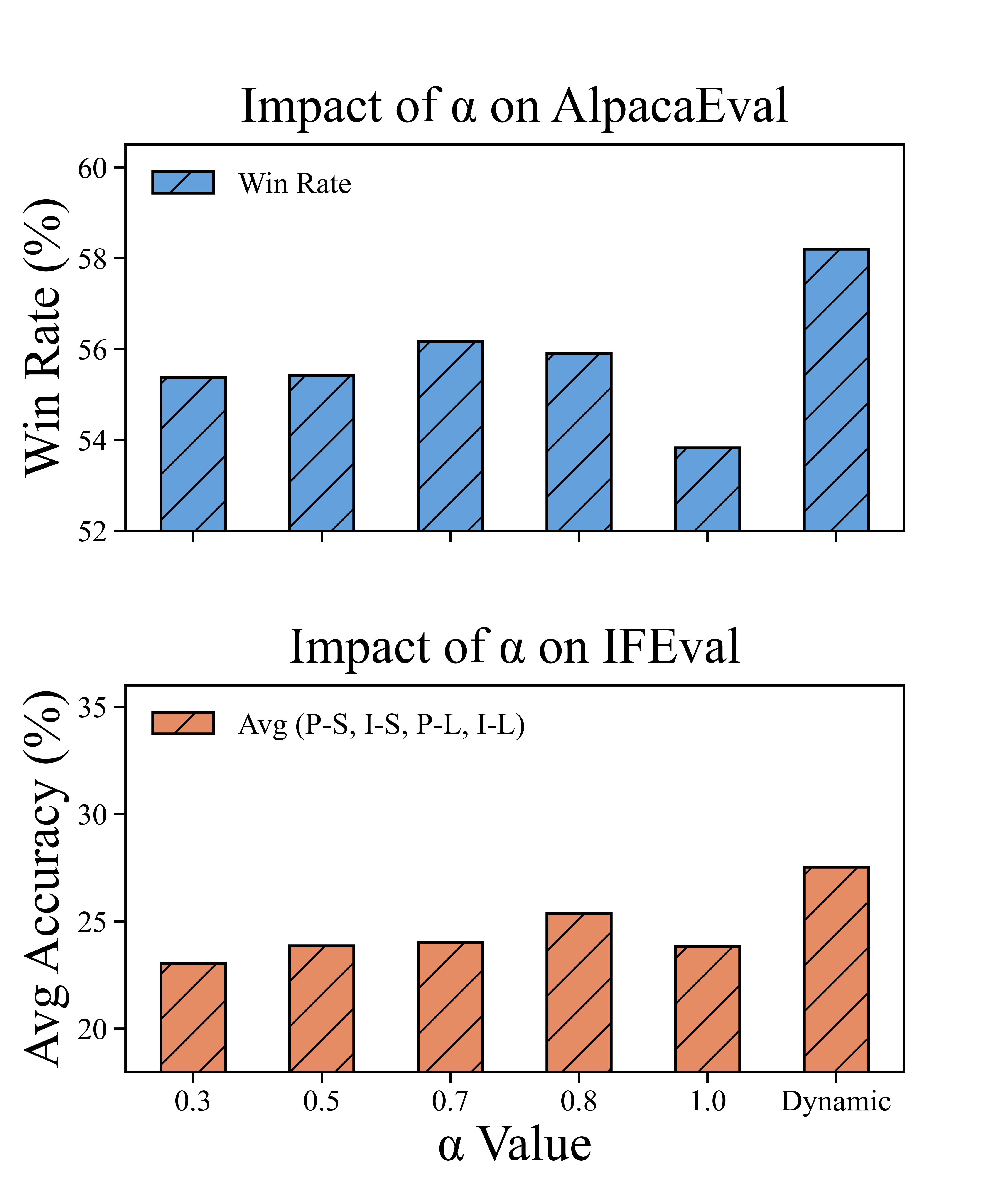}
    \caption{Evaluation of dynamic weighting strategies on LLaMA-2-7B training, comparing fixed and adaptive \(\alpha\) values using the Falcon-RefinedWeb dataset, with performance assessed on AlpacaEval and IFEval.}
    \label{fig:weighting_ablation}
\end{figure}

As shown in Figure \ref{fig:weighting_ablation}, increasing \(\alpha\) generally improves instruction-following ability and output preference by emphasizing well-aligned pairs. However, excessively high \(\alpha\) makes the model overly reliant on generated instruction-response pairs, leading to unstable training and degraded performance.

To mitigate this, our dynamic weighting strategy adaptively balances aligned and seed data, preventing instability while maintaining strong alignment. The results show that this approach significantly improves output preference and instruction-following.

\section{Multilingual Generalization}
\label{appendix:Multilingual_Generalization}
To evaluate multilingual robustness of MAIN, we perform experiments on diverse multilingual datasets.

\paragraph{Setup.}
The backbone model is Qwen-2.5-14B-Base. 
Seed data contains 3.2k instruction–response pairs from the multilingual portion of OpenAssistant, evenly sampled across French (fr), Japanese (ja), Spanish (es), Arabic (ar), and Chinese (zh). 
Additionally, 500k responses are drawn from the multilingual mC4 corpus, a naturally web collection. 
After applying mutual alignment and filtering, we obtain a balanced set of 20k pairs for supervised fine-tuning. 
Baselines use identical data and training settings to ensure comparability.

\paragraph{Benchmarks.}
Models are evaluated on multilingual reasoning and instruction-following benchmarks:
\begin{itemize}
    \item JMMLU (ja), AMMLU (ar), CMMLU (zh): 5-shot reasoning dataset. 
    \item mIFEval (fr/ja/es): multilingual instruction-following dataset.
\end{itemize}

\paragraph{Results.}
Table~\ref{tab:multi-results} summarizes multilingual evaluation on instruction-following and reasoning benchmarks . MAIN attains the best performance on these benchmarks, with the largest gains on mIFEval (fr/ja/es) and notable improvements on JMMLU and CMMLU; performance on AMMLU is on par with the strongest baseline. These results demonstrate that our framework exhibits strong generalization ability across diverse multilingual settings.

\begin{table}[h]
\centering
\small
\setlength{\tabcolsep}{2pt} 
\begin{tabular}{lccc|ccc}
\toprule
Method & \multicolumn{3}{c|}{mIFEval} & \multicolumn{3}{c}{Reasoning} \\
       & fr & ja & es & JMMLU & AMMLU & CMMLU \\
\midrule
Humpback         & 63.3 & 46.0 & 42.0 & 60.5 & 57.3 & 81.9 \\
Better Align & 65.3 & 50.1 & 43.2 & 62.7 & 59.8 & 80.6 \\
MAIN             & 69.7 & 51.8 & 47.1 & 64.9 & 59.6 & 83.5 \\
\bottomrule
\end{tabular}
\caption{Evaluation of multilingual instruction-following and reasoning dataset.}
\label{tab:multi-results}
\end{table}

\section{Detailed Benchmark Settings} 

\begin{table}[htbp]
  \centering
  \resizebox{\columnwidth}{!}{%
    \setlength{\tabcolsep}{8pt}
    \begin{tabular}{lcc}
      \toprule
      \textbf{Dataset}     & \textbf{Metric}     & \textbf{Number of Shots} \\
      \midrule
      ARC\_C               & Acc\_norm           & 25 \\
      TruthfulQA           & Mc2                 & Zero-shot \\
      Winogrande           & Acc                 & 5 \\
      HellaSwag            & Acc                 & 10 \\
      MMLU                 & Acc\_norm           & 5 \\
      AlpacaEval           & Win\_rate           & Zero-shot \\
      IFEval               & P-S, etc.           & Zero-shot \\
      \bottomrule
    \end{tabular}
  }
  \caption{Evaluation settings and key metrics for benchmark datasets under few-shot and zero-shot conditions.}
  \label{tab:benchmark_fs_zs_results}
\end{table}

We have evaluated our method under both few-shot and zero-shot conditions. Specifically, tasks such as ARC\_C, HellaSwag, Winogrande, and MMLU were tested with a few-shot setup, whereas AlpacaEval, TruthfulQA, and IFEval benchmarks were evaluated under zero-shot conditions. Detailed settings and results are provided in Table~\ref{tab:benchmark_fs_zs_results}

\section{Mutual Filtering on MMLU}
\label{appendix:mmlu_category_filter}

We compare \textsc{LLaMA-2-7B} trained with mutual alignment only (\emph{No-filter}) and with mutual alignment plus mutual filtering (\emph{Filtered}); all other settings are identical. Table~\ref{tab:mmlu_category_table} reports accuracy (\%) by MMLU super-category.

\begin{table}[h]
\centering
\footnotesize
\setlength{\tabcolsep}{4pt}
\renewcommand{\arraystretch}{1.1}
\begin{tabular}{lrrr}
\toprule
MMLU Super-Category & No-filter & Filtered & $\Delta$ \\
\midrule
STEM                & 42.7 & 45.5 & +2.8 \\
Humanities          & 45.1 & 46.3 & +1.2 \\
Social Sciences     & 46.7 & 47.2 & +0.5 \\
Other Professions   & 41.9 & 42.6 & +0.7 \\
\midrule
Overall             & 44.1 & 45.4 & +1.3 \\
\bottomrule
\end{tabular}
\caption{MMLU accuracy (\%) by super-category for \textsc{LLaMA-2-7B}. Both settings include mutual alignment; \emph{Filtered} additionally applies mutual filtering when selecting pseudo-labeled pairs.}
\label{tab:mmlu_category_table}
\end{table}

The largest gains arise in STEM, where instructions and responses tend to be tightly coupled and logically recoverable; such pairs pass bidirectional consistency checks at higher rates, yielding larger improvements. In more open-ended areas (e.g., Humanities, Social Sciences), valid responses are more diverse, so exact reconstruction is harder and gains are accordingly smaller. Overall, these deltas indicate that mutual filtering is most beneficial for categories with higher instruction–response determinism, while still providing modest positive effects elsewhere.

\section{Computational Cost Analysis} 
This section details the computational demands of our proposed method MAIN. While MAIN involves an iterative training process, the overall compute cost is carefully managed to remain modest. In each iteration, only one model is trained for a single epoch on the seed data, while the other performs inference—a lightweight operation. Typically, three iterations suffice for convergence, resulting in six training epochs across the forward and reverse models. In comparison, baseline methods \citep{li2023self,nguyen2024better} also train a reverse model and a ranking model for a similar number of epochs. We summarize estimated GPU hours in Table~\ref{tab:appendix_cost_comparison}.

\begin{table}[htbp]
  \centering
  \resizebox{\columnwidth}{!}{%
    \setlength{\tabcolsep}{4pt}
    \begin{tabular}{@{}l p{2.6cm} c c c@{}}
      \toprule
      \textbf{Method} & \textbf{Models Trained}  & \shortstack{\textbf{Inference Needed}} & \shortstack{\textbf{GPU Hours}} \\
      \midrule
      Baseline      & Ranking model + Reverse model  & No  & 5.0 \\
      \addlinespace 
      MAIN (Ours)   & Forward model + Reverse model  & Yes & 5.3 \\
      \bottomrule
    \end{tabular}%
  }
  \caption{Estimated computational cost comparison for MAIN against baselines.}
  \label{tab:appendix_cost_comparison} 

\end{table}

\section{GPT-4 Evaluation Prompt}
\label{appendix:gpt4-prompt}

The following table presents the exact prompt used to instruct GPT-4 during blind pairwise comparisons:

\begin{table}[htbp]
\centering
\renewcommand{\arraystretch}{1.2}
\begin{tabular}{p{0.95\linewidth}}
\toprule
GPT-4 Pairwise Evaluation Prompt \\
\midrule
Please act as an expert evaluator of instruction-response alignment. 
You are given a Response and two candidate instructions: Instruction A and Instruction B. 
Your task is to decide which instruction is better aligned with the response.

Evaluate based on the following aspects:

- Alignment between instruction and response \\ - logical consistency \\ - natural language fluency \\

Response:\\
\{response\}\ \\[0.5em]

Instruction A:\\
\{instruction\_A\}\ \\[0.5em]

Instruction B:\\
\{instruction\_B\} \\[0.5em]

Please output only one of the following: \\
A win — if Instruction A is clearly better aligned with the response. \\
B win — if Instruction B is clearly better aligned with the response. \\
Tie — if both instructions are equally good or equally poor. \\
\hline
\end{tabular}
\caption{Prompt template used in GPT-4 pairwise evaluation of instruction-response alignment.}
\end{table}

\end{document}